\documentclass{article}

\usepackage{arxiv}

\usepackage[utf8]{inputenc} 
\usepackage[T1]{fontenc}    
\usepackage{hyperref}       
\usepackage{url}            
\usepackage{booktabs}       
\usepackage{amsfonts}       
\usepackage{nicefrac}       
\usepackage{microtype}      
\usepackage{lipsum}		
\usepackage{graphicx}
\usepackage{natbib}
\usepackage{doi}
\usepackage{amsmath}
\usepackage{tikz}
\usepackage{geometry}
\usepackage{tikz}

\usetikzlibrary{mindmap, shadows, positioning, backgrounds}
\usetikzlibrary{positioning, shapes, arrows.meta}
\usepackage{amsmath, amssymb}

\title{YOLO26: Key Architectural Enhancements and Performance Benchmarking for Real-Time Object Detection}

\author{
  \href{https://orcid.org/0000-0002-5417-6744}{\includegraphics[scale=0.06]{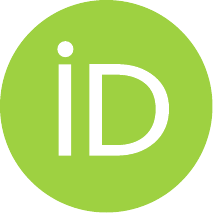}\hspace{1mm}Ranjan Sapkota\textsuperscript{1}} \quad
  Rahul Harsha Cheppally\textsuperscript{2} \quad
  Ajay Sharda\textsuperscript{2} \quad
  \href{https://orcid.org/0000-0001-5337-4848}{\includegraphics[scale=0.06]{orcid.pdf}\hspace{1mm}Manoj Karkee\textsuperscript{1}} \\
  \textsuperscript{1}Cornell University, Biological \& Environmental Engineering, Ithaca, NY 14850, USA \\
  \texttt{rs2672@cornell.edu, mk2684@cornell.edu} \\
  \textsuperscript{2}Kansas State University, Department of Biological and Agricultural Engineering, Manhattan, KS 66502, USA
}

\renewcommand{\shorttitle}{YOLO26:  (\href{https://docs.ultralytics.com/models/yolo26/ }{Ultralytics YOLO26 Official Source Link})}

\hypersetup{
pdftitle={A template for the arxiv style},
pdfsubject={q-bio.NC, q-bio.QM},
pdfauthor={David S.~Hippocampus, Elias D.~Striatum},
pdfkeywords={First keyword, Second keyword, More},
}

\begin{document}
\maketitle

\begin{abstract}
This study presents a comprehensive analysis of Ultralytics YOLO26, highlighting its key architectural enhancements and performance benchmarking for real-time edge object detection. YOLO26, released in September 2025, stands as the newest and most advanced member of the YOLO family, purpose-built to deliver efficiency, accuracy, and deployment readiness on edge and low-power devices. The paper sequentially details YOLO26’s architectural innovations, including the removal of Distribution Focal Loss (DFL), adoption of end-to-end NMS-free inference, integration of ProgLoss and Small-Target-Aware Label Assignment (STAL), and the introduction of the MuSGD optimizer for stable convergence. Beyond architecture, the study positions YOLO26 as a multi-task framework, supporting object detection, instance segmentation, pose/keypoints estimation, oriented detection, and classification. We present performance benchmarks of YOLO26 on edge devices such as NVIDIA Jetson Nano and Orin, comparing its results with YOLOv8, YOLOv11, YOLOv12, YOLOv13, and transformer-based detectors. This paper further explores real-time deployment pathways, flexible export options (ONNX, TensorRT, CoreML, TFLite), and quantization for INT8/FP16. Practical use cases of YOLO26 across robotics, manufacturing, and IoT are highlighted to demonstrate cross-industry adaptability. Finally, insights on deployment efficiency and broader implications are discussed, with future directions for YOLO26 and the YOLO lineage outlined.
\end{abstract}

\section{Introduction}
Object detection has emerged as one of the most critical tasks in computer vision, enabling machines to localize and classify multiple objects within an image or video stream \cite{zhao2019object, zou2023object}. From autonomous driving and robotics to surveillance, medical imaging, agriculture, and smart manufacturing, real-time object detection algorithms serve as the backbone of artificial intelligence (AI) applications \cite{rana2024artificial, khan2025objectdetection}. Among these algorithms, the You Only Look Once (YOLO) family has established itself as the most influential series of models for real-time object detection, combining accuracy with unprecedented inference speed \cite{sapkota2025yolo, sapkota2025rf, sapkota2024comparing, sapkota2024comparing}. Since its introduction in 2016, YOLO has evolved through numerous architectural revisions, each addressing limitations of its predecessors while integrating cutting-edge advances in neural network design, loss functions, and deployment efficiency \cite{sapkota2025yolo}. 

\begin{figure}[h!]
     \centering
     \includegraphics[width=0.79\linewidth]{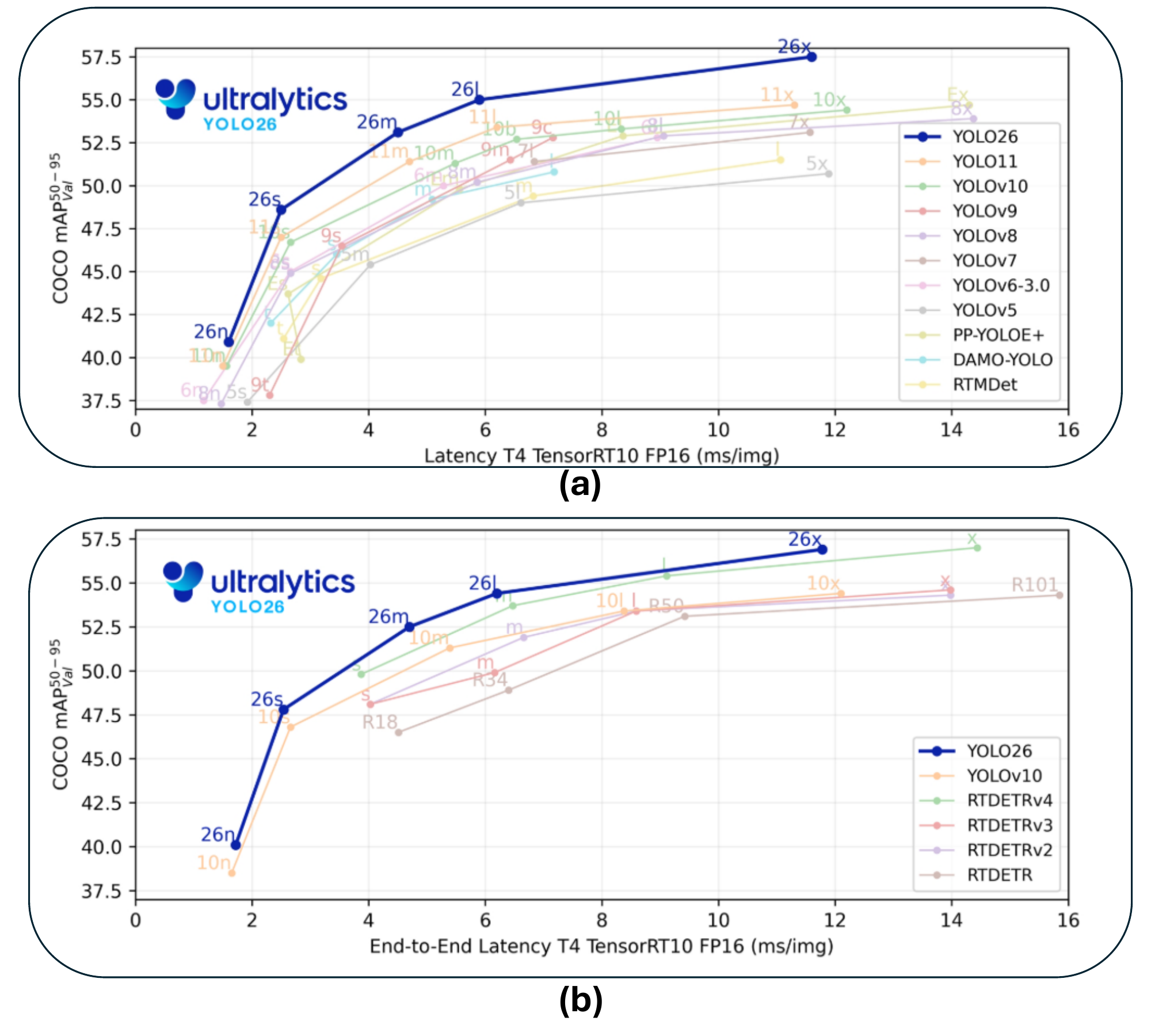}
    \caption{Performance comparison of YOLO26 under TensorRT FP16 on an NVIDIA T4 GPU  (\href{https://docs.ultralytics.com/models/yolo26/}{Source Link}). (a) COCO mAP(50–95) versus inference latency (ms/image), comparing YOLO26 with earlier YOLO versions and other real-time detectors, highlighting its improved accuracy–speed trade-off. (b) COCO mAP(50–95) versus end-to-end latency, comparing YOLO26 with YOLOv10 and RT-DETR variants, illustrating its advantage in overall pipeline efficiency.}
    \label{fig:yolograph}
\end{figure}

The release of YOLO26 in September 2025 marks the newest milestone in the YOLO lineage, shifting the design emphasis from incremental architectural complexity toward deployment-oriented simplification—most notably through streamlined regression, end-to-end prediction behavior, and training-time refinements enabled by novel optimization. This edge-first philosophy is reflected in the comparative accuracy–latency trends shown in Fig.~\ref{fig:yolograph}a, where Ultralytics reports YOLO26’s COCO mAP(50-95) versus latency performance (T4, TensorRT10, FP16) against a broad set of prior YOLO variants (YOLO11, YOLOv10, YOLOv9, YOLOv8, YOLOv7, YOLOv6-3.0, YOLOv5) as well as competitive real-time detectors (PP-YOLOE+, DAMO-YOLO, and RTMDet). Complementing this, Fig.~\ref{fig:yolograph}b positions YOLO26 on the same COCO mAP(50-95) versus end-to-end latency axis against transformer-style real-time baselines (YOLOv10 and the RT-DETR family), underscoring that YOLO26 aims to retain high detection quality while reducing overall pipeline delay, an especially relevant trade-off for low-power and latency-sensitive edge devices.

Table \ref{tab:comparison} provides a detailed comparison of YOLO models from version YOLOv1 to YOLOv13 and YOLO26, highlighting their release years, key architectural innovations, performance enhancements, and development frameworks.

\begin{table}[h!]\centering
\caption{Summary of YOLOv1 to YOLOv13 and YOLOv26 models: release year, architecture, innovations, frameworks}
\label{tab:comparison}
\scriptsize
\begin{tabular}{p{4cm} p{6.5cm} p{2.2cm} p{2cm}}
\hline
\textbf{Model (Year)} & \textbf{Key Architectural Innovation and Contribution} & \textbf{Tasks} & \textbf{Framework} \\
\hline
YOLOv1 (2015) \cite{redmon2016you} & First unified single-stage object detector (one network for bounding boxes + class probabilities). & Object Detection, Classification & Darknet \\
YOLOv2 (2016)  \cite{redmon2017yolo9000} & Multi-scale training introduced; anchor box dimension clustering for improved prior boxes (YOLO9000 joint detection/classification). & Object Detection, Classification & Darknet \\
YOLOv3 (2018) \cite{redmon2018yolov3} & Deeper Darknet-53 backbone with residual connections; added SPP module and multi-scale feature fusion for small object detection. & Object Detection, Multi-scale Detection & Darknet \\
YOLOv4 (2020) \cite{bochkovskiy2020yolov4} & Mish activation function adopted; CSPDarknet-53 backbone (Cross-Stage Partial networks) for enhanced feature reuse. & Object Detection, Object Tracking & Darknet \\
YOLOv5 (2020) (\href{https://github.com/ultralytics/yolov5}{Source Link}) & PyTorch implementation by Ultralytics; anchor-free detection head option; used SiLU (Swish) activation and PANet neck for feature aggregation. & Object Detection, Instance Segmentation (limited) & PyTorch \newline  (Ultralytics) \\
YOLOv6 (2022) \cite{li2022yolov6} & EfficientRep backbone with embedded self-attention; introduced anchor-free object detection mode for efficiency. & Object Detection, Instance Segmentation & PyTorch \\
YOLOv7 (2022) \cite{wang2023yolov7} & Extended ELAN (E-ELAN) backbone with model re-parameterization; integrated transformer-based modules for broader tasks (e.g. tracking). & Object Detection, Object Tracking, Instance Segmentation & PyTorch \\
YOLOv8 (2023) (\href{https://docs.ultralytics.com/models/yolov8/}{Source Link}) & Ultralytics next-gen model; new C2f backbone and decoupled head; incorporated generative techniques (GAN-based augmentation) and fully anchor-free design. & Object Detection, Instance Segmentation, Panoptic Segmentation, Keypoint Estimation & PyTorch \newline (Ultralytics) \\
YOLOv9 (2024) \cite{wang2024yolov9} & Introduced Programmable Gradient Information (PGI) for selective learning; proposed G-ELAN (an enhanced ELAN architecture) for improved feature extraction. & Object Detection, Instance Segmentation & PyTorch \\
YOLOv10 (2024) \cite{wang2024yolov10} & Achieved end-to-end NMS-free detection via a consistent dual-assignment training strategy (removing post-processing). & Object Detection & PyTorch \\
YOLO11 (2024) (\href{https://docs.ultralytics.com/models/yolo11/}{Source Link}) & Added C3k2 CSP bottlenecks (smaller kernel CSP blocks) throughout backbone/neck for efficiency; retained SPPF and introduced C2PSA (CSP with Spatial Attention) module to focus on important regions. Extended YOLO to pose estimation and oriented object detection tasks. & Object Detection, Instance Segmentation, Pose Estimation, Oriented Detection & PyTorch \newline (Ultralytics) \\
YOLOv12 (2025) \cite{tian2025yolov12} & Attention-centric architecture: introduced an efficient area attention module (global self-attention with low complexity) and Residual ELAN (R-ELAN) blocks to improve feature aggregation, achieving transformer-level accuracy at YOLO speeds. & Object Detection & PyTorch \\
YOLOv13 (2025) \cite{lei2025yolov13} & Hypergraph-based Adaptive Correlation Enhancement (HyperACE) module to capture global high-order feature interactions; Full-Pipeline Aggregation-Distribution (FullPAD) scheme for enhanced feature flow throughout the network; utilized depthwise-separable convolutions to reduce complexity. & Object Detection & PyTorch \\
YOLOv26 (2025) (\href{https://docs.ultralytics.com/models/yolo26/}{Source Link}) & Ultralytics edge-optimized model: eliminated NMS with a native end-to-end predictor; removed DFL (Distribution Focal Loss) for simpler, faster inference; introduced MuSGD optimizer (SGD+Muon hybrid) for stable and quick convergence; significantly improved small-object accuracy and up to 43\% faster CPU inference for deployment on low-power devices. & Object Detection, Instance Segmentation, Pose Estimation, Oriented Detection, Classification & PyTorch \newline(Ultralytics) \\
\hline
\end{tabular}
\end{table}

The YOLO framework was first proposed by Joseph Redmon and colleagues in 2016, introducing a paradigm shift in object detection \cite{redmon2016you}. Unlike traditional two-stage detectors such as R-CNN \cite{he2017mask} and Faster R-CNN \cite{ren2016faster}, which separated region proposal from classification, YOLO formulated detection as a single regression problem \cite{diwan2023object}. By directly predicting bounding boxes and class probabilities in one forward pass through a convolutional neural network (CNN), YOLO achieved real-time speeds while maintaining competitive accuracy \cite{ali2024yolo, diwan2023object}. This efficiency made YOLOv1 highly attractive for applications where latency was a critical factor, including robotics, autonomous navigation, and live video analytics. Subsequent versions YOLOv2 (2017)  \cite{redmon2017yolo9000}and YOLOv3 (2018) \cite{redmon2018yolov3} significantly improved accuracy while retaining real-time performance. YOLOv2 introduced batch normalization, anchor boxes, and multi-scale training, which increased robustness across varying object sizes. YOLOv3 leveraged a deeper architecture based on Darknet-53, along with multi-scale feature maps for better small-object detection. These enhancements made YOLOv3 the de facto standard for academic and industrial applications for several years \cite{apostolidis2025delving, sapkota2025yolo, edozie2025comprehensive}. 

As the demand for higher accuracy grew, especially in challenging domains such as aerial imagery, agriculture, and medical analysis, YOLO models diversified into more advanced architectures. YOLOv4 (2020) \cite{bochkovskiy2020yolov4} introduced Cross-Stage Partial Networks (CSPNet), improved activation functions like Mish, and advanced training strategies including mosaic data augmentation and CIoU loss. YOLOv5 (Ultralytics, 2020), though unofficial, gained immense popularity due to its PyTorch implementation, extensive community support, and simplified deployment across diverse platforms. YOLOv5 also brought modularity, making it easier to adapt for segmentation, classification, and edge applications. Further developments included YOLOv6\cite{li2022yolov6} and YOLOv7 \cite{wang2023yolov7} (2022), which integrated advanced optimization techniques, parameter-efficient modules, and transformer-inspired blocks. These iterations pushed YOLO closer to state-of-the-art (SoTA) accuracy benchmarks while retaining a focus on real-time inference. The YOLO ecosystem, by this point, had firmly established itself as the leading family of models in object detection research and deployment. 

Ultralytics, the primary maintainer of modern YOLO releases, redefined the framework with YOLOv8 (2023) \cite{sohan2024review}. YOLOv8 featured a decoupled detection head, anchor-free predictions, and refined training strategies, resulting in substantial improvements in both accuracy and deployment versatility \cite{farooq2024improved}. It was widely adopted in industry due to its clean Python API, compatibility with TensorRT, CoreML, and ONNX, and availability of variants optimized for speed versus accuracy trade-offs (nano, small, medium, large, and extra-large). YOLOv9 \cite{wang2024yolov9}, YOLOv10 \cite{wang2024yolov10}, and YOLO11 followed in rapid succession, each iteration pushing the boundaries of architecture and performance. YOLOv9 introduced GELAN (Generalized Efficient Layer Aggregation Network) and Progressive Distillation, combining efficiency with higher representational capacity. YOLOv10 focused on balancing accuracy and inference latency with hybrid task-aligned assignments. YOLOv11 further refined Ultralytics’ vision, offering higher efficiency on GPUs while maintaining strong small-object performance \cite{sapkota2025yolo}. Together, these models cemented Ultralytics’ reputation for producing production-ready YOLO releases tailored to modern deployment pipelines.

Following YOLO11, alternative versions YOLOv12\cite{tian2025yolov12} and YOLOv13 \cite{lei2025yolov13} introduced attention-centric designs and advanced architectural components that sought to maximize accuracy across diverse datasets. These models explored multi-head self-attention, improved multi-scale fusion, and stronger training regularization strategies. While they offered strong benchmarks, they retained reliance on Non-Maximum Suppression (NMS) and Distribution Focal Loss (DFL), which introduced latency overhead and export challenges, especially for low-power devices. The limitations of NMS-based post-processing and complex loss formulations motivated the development of YOLO26  (\href{https://docs.ultralytics.com/models/yolo26/}{Ultralytics YOLO26 Official Source}). By September 2025, at the YOLO Vision 2025 event in London, Ultralytics unveiled YOLO26 as a next-generation model optimized for edge computing, robotics, and mobile AI. 

\begin{figure}[h!]
     \centering
     \includegraphics[width=0.66\linewidth]{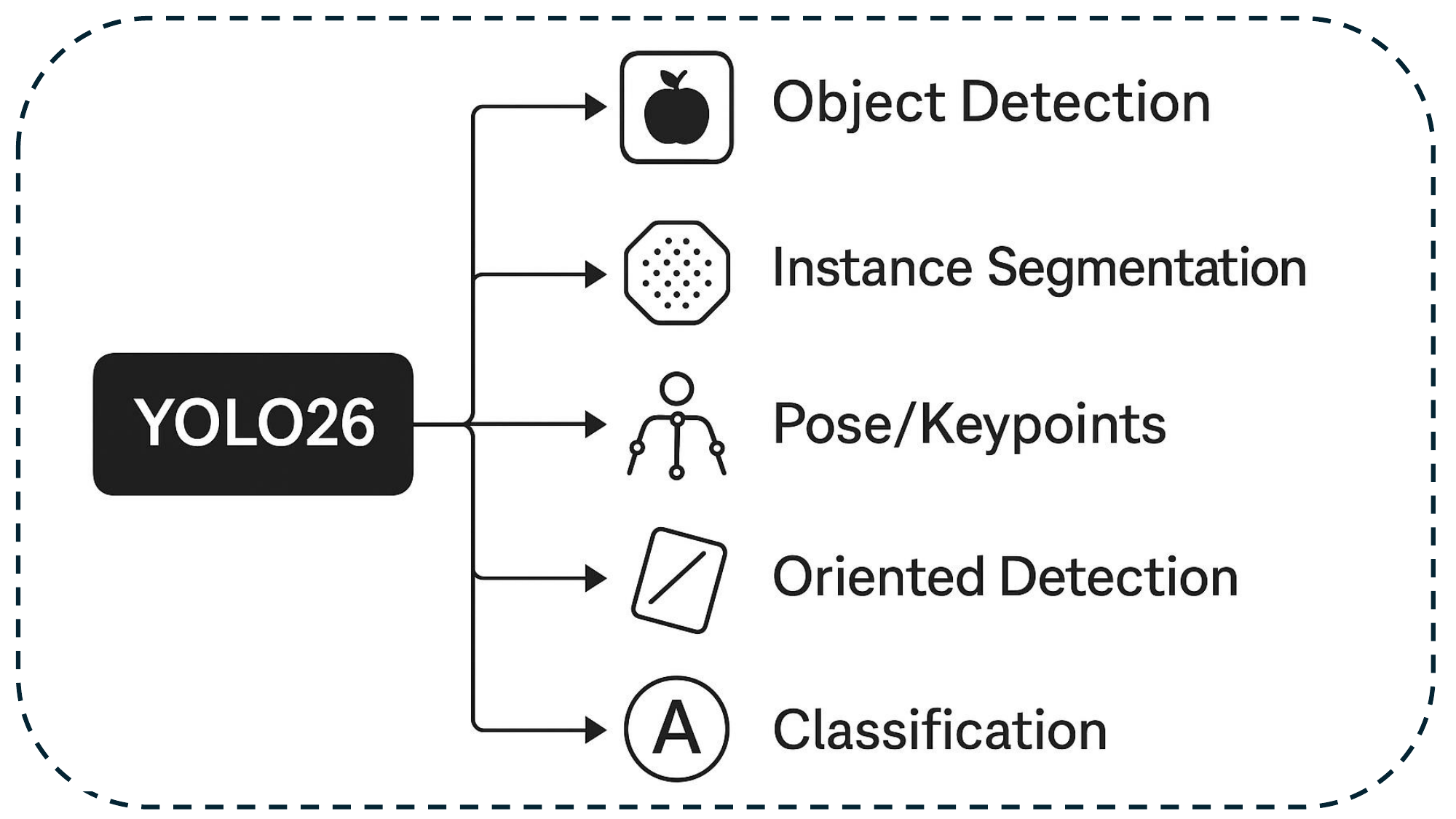}
    \caption{YOLO26 unified architecture supports five key vision tasks object detection, instance segmentation, pose/keypoints detection, oriented detection, and classification.}
    \label{fig:yolointro}
\end{figure}

YOLO26 is engineered around three guiding principles simplicity, efficiency, and innovation and the overview in Figure \ref{fig:yolointro} situates these choices alongside its five supported tasks: object detection, instance segmentation, pose/keypoints detection, oriented detection, and classification. On the inference path, YOLO26 eliminates NMS, producing native end-to-end predictions that remove a major post-processing bottleneck, reduce latency variance, and simplify threshold tuning across deployments. On the regression side, it removes DFL, turning distributional box decoding into a lighter, hardware-friendly formulation that exports cleanly to ONNX, TensorRT, CoreML, and TFLite a practical win for edge and mobile pipelines. Together, these changes yield a leaner graph, faster cold-start, and fewer runtime dependencies, which is particularly beneficial for CPU-bound and embedded scenarios. Training stability and small-object fidelity are addressed through ProgLoss (progressive loss balancing) and STAL (small-target-aware label assignment). ProgLoss adaptively reweights objectives to prevent domination by easy examples late in training, while STAL prioritizes assign- ment for tiny or occluded instances, improving recall under clutter, foliage, or motion blur conditions common in aerial, robotics, and smart-camera feeds. Optimization is driven by MuSGD, a hybrid that blends the generalization of SGD with momentum/curvature behaviors inspired by Muon-style methods, enabling faster, smoother convergence and more reliable plateaus across scales.

Functionally, as highlighted again in Figure \ref{fig:yolointro}, YOLO26’s five capabilities share a unified backbone/neck and streamlined heads:
\begin{itemize}
    \item \textbf{Object Detection:} Anchor-free, NMS-free boxes and scores
    \item \textbf{Instance Segmentation:} Lightweight mask branches coupled to shared features;
    \item \textbf{Pose/Keypoints Detection:} Compact keypoint heads for human or part landmarks
    \item \textbf{Oriented Detection:} Rotated boxes for oblique objects and elongated targets
    \item \textbf{Classification:} Single-label logits for pure recognition tasks.
\end{itemize}

This consolidated design allows multi-task training or task-specific fine-tuning without architectural rework, while the simplified exports preserve portability across accelerators. In sum, YOLO26 advances the YOLO lineage by pairing end-to-end inference and DFL-free regression with ProgLoss, STAL, and MuSGD, yielding a model that is faster to deploy, steadier to train, and broader in capability as visually summarized in Figure \ref{fig:yolointro}.

\section{Architectural Enhancements in YOLO26}
\begin{figure}[ht!]
     \centering
     \includegraphics[width=0.90\linewidth]{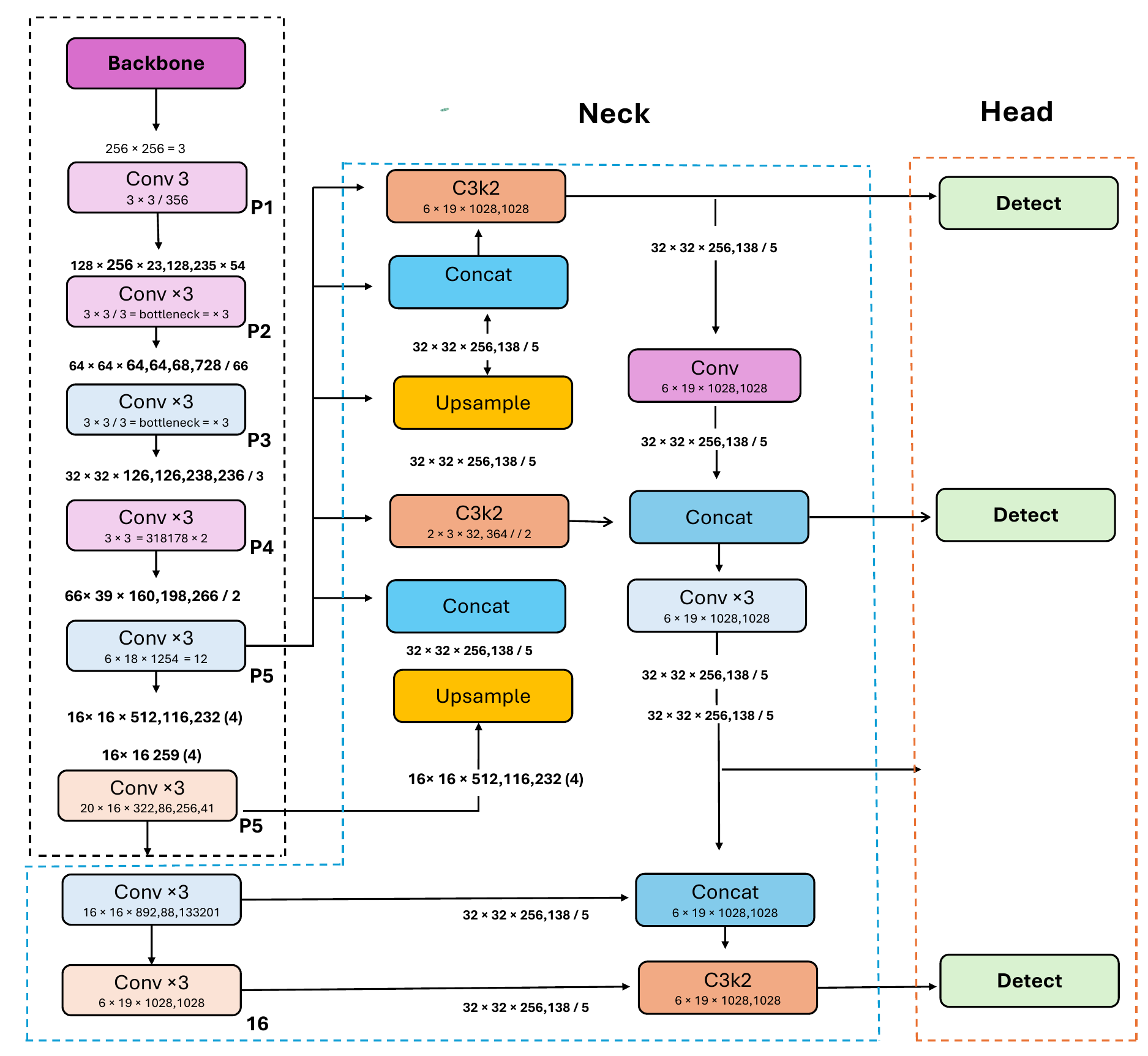}
    \caption{Core architecture diagram of Ultralytics YOLO26 object detection and segmentation algorithm}
    \label{fig:simplifiedarchitecture}
\end{figure}

The architecture of YOLO26 follows a streamlined and efficient pipeline that has been purpose-built for real-time object detection across edge and server platforms. As illustrated in Figure \ref{fig:simplifiedarchitecture}, the process begins with the ingestion of input data in the form of images or video streams, which are first passed through preprocessing operations including resizing and normalization to standard dimensions suitable for model inference. The data is then fed into the backbone feature extraction stage, where a compact yet powerful convolutional network captures hierarchical representations of visual patterns. To enhance robustness across scales, the architecture generates multi-scale feature maps (Figure \ref{fig:simplifiedarchitecture}) that preserve semantic richness for both large and small objects. These feature maps are then merged within a lightweight feature fusion neck, where information is integrated in a computationally efficient manner. Detection-specific processing occurs in the direct regression head, which, unlike prior YOLO versions, outputs bounding boxes and class probabilities without relying on Non-Maximum Suppression (NMS). This end-to-end NMS-free inference (Figure \ref{fig:simplifiedarchitecture}) eliminates post-processing overhead and accelerates deployment. Training stability and accuracy are reinforced by ProgLoss balancing and STAL assignment modules, which ensure equitable weighting of loss terms and improved detection of small targets. Model optimization is guided by the MuSGD optimizer, combining the strengths of SGD and Muon for faster and more reliable convergence. Deployment efficiency is further enhanced through quantization, with support for FP16 and INT8 precision, enabling acceleration on CPUs, NPUs, and GPUs with minimal accuracy degradation. Finally, the pipeline culminates in the generation of output predictions, including bounding boxes and class assignments that can be visualized overlaid on the input image. Overall, the architecture of YOLO26 demonstrates a carefully balanced design philosophy that simultaneously advances accuracy, stability, and deployment simplicity.

\begin{figure}[ht!]
     \centering
     \includegraphics[width=0.70\linewidth]{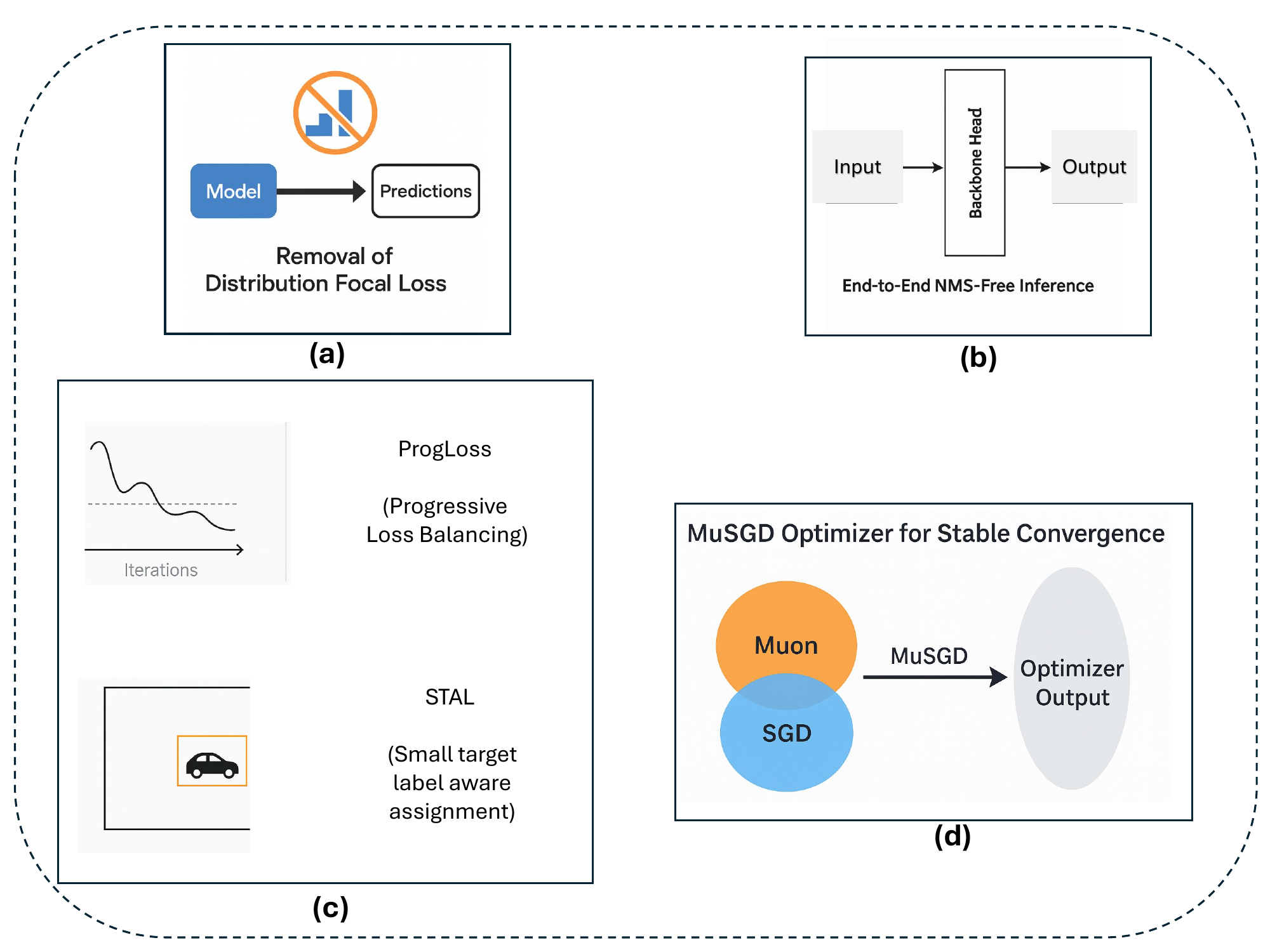}
    \caption{Key architectural enhancements in YOLO26: (a) Removal of Distribution Focal Loss (DFL) streamlines bounding box regression, boosting efficiency and export compatibility. (b) End-to-end NMS-free inference eliminates post-processing bottlenecks, enabling faster and simpler deployment. (c) ProgLoss and STAL enhance training stability and significantly improve small-object detection accuracy. (d) The MuSGD optimizer combines SGD and Muon strengths, achieving faster, more stable convergence in training. }
    \label{fig:architectures}
\end{figure}

YOLO26 introduces several key architectural innovations that differentiate it from prior generations of YOLO models. These enhancements not only improve training stability and inference efficiency but also fundamentally reshape the deployment pipeline for real-time edge devices. In this section, we describe four major contributions of YOLO26: (i) the removal of Distribution Focal Loss (DFL), (ii) the introduction of end-to-end Non-Maximum Suppression (NMS)-free inference, (iii) novel loss function strategies including Progressive Loss Balancing (ProgLoss) and Small-Target-Aware Label Assignment (STAL), and (iv) the development of the MuSGD optimizer for stable and efficient convergence. Each of these architectural enhancements is discussed in detail, with comparative insights highlighting their advantages over earlier YOLO versions such as YOLOv8, YOLOv11, YOLOv12, and YOLOv13.

\subsection{Removal of Distribution Focal Loss (DFL)}

One of the most significant architectural simplifications in YOLO26 is the removal of the Distribution Focal Loss (DFL) module (Figure \ref{fig:architectures}a), which had been present in prior YOLO releases such as YOLOv8 and YOLOv11. DFL was originally designed to improve bounding box regression by predicting probability distributions for box coordinates, thereby allowing more precise localization of objects. While this strategy demonstrated accuracy gains in earlier models, it also introduced non-trivial computational overhead and export difficulties. In practice, DFL required specialized handling during inference and model export, which complicated deployment pipelines targeting hardware accelerators such as ONNX, CoreML, TensorRT, or TFLite.  

By eliminating DFL, YOLO26 simplifies the model’s architecture, making bounding box prediction a more straightforward regression task without sacrificing performance. Comparative analysis indicates that YOLO26 achieves comparable or superior accuracy to DFL-based YOLO models, particularly when combined with other innovations such as ProgLoss and STAL. Moreover, the removal of DFL significantly reduces inference latency and improves cross-platform compatibility. This makes YOLO26 more suitable for edge AI scenarios, where lightweight and hardware-friendly models are paramount.  

In contrast, models such as YOLOv12 and YOLOv13 retained DFL in their architectures, which limited their applicability on constrained devices despite strong accuracy benchmarks on GPU-rich environments. YOLO26 therefore marks a decisive step toward aligning state-of-the-art object detection performance with the realities of mobile, embedded, and industrial applications.

\subsection{End-to-End NMS-Free Inference}

Another groundbreaking feature of YOLO26 is its native support for end-to-end inference without Non-Maximum Suppression (NMS) (Refer to Figure \ref{fig:architectures}b). Traditional YOLO models, including YOLOv8 through YOLOv13, rely heavily on NMS as a post-processing step to filter out duplicate predictions by retaining only the bounding boxes with the highest confidence scores. While effective, NMS adds additional latency to the pipeline and requires manually tuned hyperparameters such as the Intersection-over-Union (IoU) threshold. This dependence on a handcrafted post-processing step introduces fragility in deployment pipelines, especially for edge devices and latency-sensitive applications.  

YOLO26 fundamentally redesigns the prediction head to produce direct, non-redundant bounding box predictions without the need for NMS. This end-to-end design not only reduces inference complexity but also eliminates the dependency on hand-tuned thresholds, thereby simplifying integration into production systems. Comparative benchmarks demonstrate that YOLO26 achieves faster inference speeds than YOLOv11 and YOLOv12, with CPU inference times reduced by up to 43\% for the nano model. This makes YOLO26 particularly advantageous for mobile devices, UAVs, and embedded robotics platforms where milliseconds of latency can have substantial operational impacts.  

Beyond speed, the NMS-free approach improves reproducibility and deployment portability, as models no longer require extensive post-processing code. While other advanced detectors such as RT-DETR and Sparse R-CNN have experimented with NMS-free inference, YOLO26 represents the first YOLO release to adopt this paradigm while maintaining YOLO’s hallmark balance between speed and accuracy. Compared to YOLOv13, which still depends on NMS, YOLO26’s end-to-end pipeline stands out as a forward-looking architecture for real-time detection.

\subsection{ProgLoss and STAL: Enhanced Training Stability and Small-Object Detection}

Training stability and small-object recognition remain persistent challenges in object detection. YOLO26 addresses these through the integration of two novel strategies: Progressive Loss Balancing (ProgLoss) and Small-Target-Aware Label Assignment (STAL), as depicted in Figure (Figure \ref{fig:architectures}c) 

ProgLoss dynamically adjusts the weighting of different loss components during training, ensuring that the model does not overfit to dominant object categories while underperforming on rare or small classes. This progressive rebalancing improves generalization and prevents instability during later epochs of training. STAL, on the other hand, explicitly prioritizes label assignments for small objects, which are particularly difficult to detect due to their limited pixel representation and susceptibility to occlusion. Together, ProgLoss and STAL provide YOLO26 with a substantial accuracy boost on datasets with small or occluded objects, such as COCO and UAV imagery benchmarks.  

Comparatively, earlier models such as YOLOv8 and YOLOv11 did not incorporate such targeted mechanisms, often requiring dataset-specific augmentations or external training tricks to achieve acceptable small-object performance. YOLOv12 and YOLOv13 attempted to address this gap through attention-based modules and enhanced multi-scale feature fusion; however, these solutions increased architectural complexity and inference costs. YOLO26 achieves similar or superior improvements with a more lightweight approach, reinforcing its suitability for edge AI applications. By integrating ProgLoss and STAL, YOLO26 establishes itself as a robust small-object detector while maintaining the efficiency and portability of the YOLO family.

\subsection{MuSGD Optimizer for Stable Convergence}

A final innovation in YOLO26 is the introduction of the MuSGD optimizer (Figure \ref{fig:architectures}d), which combines the strengths of Stochastic Gradient Descent (SGD) with the recently proposed Muon optimizer, a technique inspired by optimization strategies used in large language model (LLM) training. MuSGD leverages the robustness and generalization capacity of SGD while incorporating adaptive properties from Muon, enabling faster convergence and more stable optimization across diverse datasets.  

This hybrid optimizer reflects an important trend in modern deep learning: the cross-pollination of advances between natural language processing (NLP) and computer vision. By borrowing from LLM training practices (e.g., Kimi K2 by Moonshot AI), YOLO26 benefits from stability enhancements that were previously unexplored in the YOLO lineage. Empirical results show that MuSGD enables YOLO26 to reach competitive accuracy with fewer training epochs, reducing both training time and computational cost.  

Previous YOLO versions, including YOLOv8 through YOLOv13, relied on standard SGD or AdamW variants. While effective, these optimizers required extensive hyperparameter tuning and sometimes exhibited unstable convergence, particularly on datasets with high variability. In comparison, MuSGD improves reliability while preserving YOLO’s lightweight training ethos. For practitioners, this translates into shorter development cycles, fewer training restarts, and more predictable performance across deployment scenarios. By integrating MuSGD, YOLO26 positions itself as not only an inference-optimized model but also a training-friendly architecture for researchers and industry practitioners alike.

\section{Benchmarking and Comparative Analysis}

\subsection{Detection and Segmentation Performance Metrics}
As summarized in the Detection results in Table~\ref{tab:yolo26_metrics}, YOLO26 consistently improves COCO mAP(50–95) as model scale increases, while maintaining predictable and low inference latency across CPU (ONNX) and GPU (TensorRT) runtimes. In particular, YOLO26-m and YOLO26-l achieve strong detection accuracy above 53\% and 55\% mAP, respectively, at substantially lower latency than transformer-based alternatives, reflecting the benefits of its NMS-free inference path and simplified regression design. The same table further highlights favorable scaling behavior in parameters and FLOPs, reinforcing YOLO26’s efficiency across deployment targets. 

Beyond detection, the Segmentation results in Table~\ref{tab:yolo26_metrics} demonstrate that YOLO26 retains these advantages in multi-task settings. Across nano to extra-large variants, YOLO26-seg models deliver competitive box and mask mAP while preserving manageable computational cost and real-time throughput, even under end-to-end evaluation. When contrasted with architectures such as YOLOv10 and RT-DETR variants, which rely on heavier transformer encoders, YOLO26 exhibits a more balanced accuracy–latency profile, particularly for edge and CPU-bound inference. Taken together, the detection and segmentation benchmarks in Table~\ref{tab:yolo26_metrics} show that YOLO26 is not merely an incremental refinement, but a deployment-oriented evolution of the YOLO family, effectively bridging efficiency-focused design and high-accuracy real-time perception under stringent latency constraints.

\begin{table*}[h!]
\centering
\scriptsize
\setlength{\tabcolsep}{4pt}
\renewcommand{\arraystretch}{1.15}
\caption{Ultralytics YOLO26 performance metrics (640 px). Detection (top) and instance segmentation (bottom) results report COCO validation accuracy, end-to-end (e2e) scores where applicable, and speed on CPU (ONNX) and NVIDIA T4 (TensorRT10 FP16), along with model size (params) and compute (FLOPs).}
\label{tab:yolo26_metrics}
\vspace{2pt}

\begin{tabular}{l c c c c c c c}
\toprule
\multicolumn{8}{l}{\textbf{Detection}} \\
\midrule
\textbf{Model} & \textbf{Size} & \textbf{mAP$_{val}$} & \textbf{mAP$_{val}$} & \textbf{Speed} & \textbf{Speed} & \textbf{Params} & \textbf{FLOPs} \\
 & \textbf{(px)} & \textbf{50--95} & \textbf{50--95 (e2e)} & \textbf{CPU ONNX (ms)} & \textbf{T4 TRT10 (ms)} & \textbf{(M)} & \textbf{(B)} \\
\midrule
YOLO26n & 640 & 40.9 & 40.1 & $38.9 \pm 0.7$ & $1.7 \pm 0.0$ & 2.4 & 5.4 \\
YOLO26s & 640 & 48.6 & 47.8 & $87.2 \pm 0.9$ & $2.5 \pm 0.0$ & 9.5 & 20.7 \\
YOLO26m & 640 & 53.1 & 52.5 & $220.0 \pm 1.4$ & $4.7 \pm 0.1$ & 20.4 & 68.2 \\
YOLO26l & 640 & 55.0 & 54.4 & $286.2 \pm 2.0$ & $6.2 \pm 0.2$ & 24.8 & 86.4 \\
YOLO26x & 640 & 57.5 & 56.9 & $525.8 \pm 4.0$ & $11.8 \pm 0.2$ & 55.7 & 193.9 \\
\bottomrule
\end{tabular}

\vspace{6pt}

\begin{tabular}{l c c c c c c c}
\toprule
\multicolumn{8}{l}{\textbf{Instance Segmentation}} \\
\midrule
\textbf{Model} & \textbf{Size} & \textbf{mAP$_{box}$} & \textbf{mAP$_{mask}$} & \textbf{Speed} & \textbf{Speed} & \textbf{Params} & \textbf{FLOPs} \\
 & \textbf{(px)} & \textbf{50--95 (e2e)} & \textbf{50--95 (e2e)} & \textbf{CPU ONNX (ms)} & \textbf{T4 TRT10 (ms)} & \textbf{(M)} & \textbf{(B)} \\
\midrule
YOLO26n-seg & 640 & 39.6 & 33.9 & $53.3 \pm 0.5$ & $2.1 \pm 0.0$ & 2.7 & 9.1 \\
YOLO26s-seg & 640 & 47.3 & 40.0 & $118.4 \pm 0.9$ & $3.3 \pm 0.0$ & 10.4 & 34.2 \\
YOLO26m-seg & 640 & 52.5 & 44.1 & $328.2 \pm 2.4$ & $6.7 \pm 0.1$ & 23.6 & 121.5 \\
YOLO26l-seg & 640 & 54.4 & 45.5 & $387.0 \pm 3.7$ & $8.0 \pm 0.1$ & 28.0 & 139.8 \\
YOLO26x-seg & 640 & 56.5 & 47.0 & $787.0 \pm 6.8$ & $16.4 \pm 0.1$ & 62.8 & 313.5 \\
\bottomrule
\end{tabular}
\end{table*}

\subsection{Classification performance metrics (ImageNet)}

Table~\ref{tab:yolo26_classification} summarizes the ImageNet classification performance of YOLO26 across model scales. As model capacity increases from YOLO26n to YOLO26x, Top-1 accuracy improves steadily from 71.4\% to 79.9\%, while maintaining strong Top-5 accuracy above 90\% for all variants. Importantly, this accuracy scaling is achieved with predictable latency growth, as TensorRT FP16 inference remains below 4\,ms even for the largest model. The compact FLOPs and parameter counts reported in Table~\ref{tab:yolo26_classification} highlight that YOLO26 classification heads preserve efficiency, making them well suited for real-time image recognition on edge and embedded platforms.

\begin{table}[t]
\centering
\scriptsize
\setlength{\tabcolsep}{4pt}
\renewcommand{\arraystretch}{1.15}
\caption{YOLO26 image classification performance on ImageNet at 224 px resolution. Results report Top-1/Top-5 accuracy, inference speed on CPU (ONNX) and NVIDIA T4 (TensorRT10 FP16), along with model size and FLOPs.}
\label{tab:yolo26_classification}
\vspace{2pt}

\begin{tabular}{l c c c c c c c}
\toprule
\textbf{Model} & \textbf{Size} & \textbf{Acc} & \textbf{Acc} & \textbf{Speed} & \textbf{Speed} & \textbf{Params} & \textbf{FLOPs} \\
 & \textbf{(px)} & \textbf{Top-1} & \textbf{Top-5} & \textbf{CPU ONNX (ms)} & \textbf{T4 TRT10 (ms)} & \textbf{(M)} & \textbf{(B @224)} \\
\midrule
YOLO26n-cls & 224 & 71.4 & 90.1 & $5.0 \pm 0.3$ & $1.1 \pm 0.0$ & 2.8 & 0.5 \\
YOLO26s-cls & 224 & 76.0 & 92.9 & $7.9 \pm 0.2$ & $1.3 \pm 0.0$ & 6.7 & 1.6 \\
YOLO26m-cls & 224 & 78.1 & 94.2 & $17.2 \pm 0.4$ & $2.0 \pm 0.0$ & 11.6 & 4.9 \\
YOLO26l-cls & 224 & 79.0 & 94.6 & $23.2 \pm 0.3$ & $2.8 \pm 0.0$ & 14.1 & 6.2 \\
YOLO26x-cls & 224 & 79.9 & 95.0 & $41.4 \pm 0.9$ & $3.8 \pm 0.0$ & 29.6 & 13.6 \\
\bottomrule
\end{tabular}
\end{table}

\subsection{Pose Performance Metrics (COCO)}
Table~\ref{tab:yolo26_pose} presents the pose estimation performance of YOLO26 on the COCO dataset. Across model scales, YOLO26 exhibits consistent gains in mAP$_{pose}$, increasing from 57.2\% for the nano variant to 71.6\% for the extra-large model under end-to-end evaluation. This accuracy improvement is accompanied by predictable scaling in latency and computational cost, while maintaining real-time inference on GPU and near-real-time performance on CPU. The results in Table~\ref{tab:yolo26_pose} demonstrate that YOLO26 effectively extends its efficiency-oriented design to pose estimation, making it suitable for real-time human and object keypoint analysis on both edge and server platforms.

\begin{table}[h!]
\centering
\scriptsize
\setlength{\tabcolsep}{4pt}
\renewcommand{\arraystretch}{1.15}
\caption{YOLO26 pose estimation performance on the COCO dataset at 640 px resolution. Results report end-to-end (e2e) pose accuracy, inference speed on CPU (ONNX) and NVIDIA T4 (TensorRT10 FP16), along with model size and FLOPs.}
\label{tab:yolo26_pose}
\vspace{2pt}

\begin{tabular}{l c c c c c c c}
\toprule
\textbf{Model} & \textbf{Size} & \textbf{mAP$_{pose}$} & \textbf{mAP$_{pose}$} & \textbf{Speed} & \textbf{Speed} & \textbf{Params} & \textbf{FLOPs} \\
 & \textbf{(px)} & \textbf{50--95 (e2e)} & \textbf{50 (e2e)} & \textbf{CPU ONNX (ms)} & \textbf{T4 TRT10 (ms)} & \textbf{(M)} & \textbf{(B)} \\
\midrule
YOLO26n-pose & 640 & 57.2 & 83.3 & $40.3 \pm 0.5$ & $1.8 \pm 0.0$ & 2.9 & 7.5 \\
YOLO26s-pose & 640 & 63.0 & 86.6 & $85.3 \pm 0.9$ & $2.7 \pm 0.0$ & 10.4 & 23.9 \\
YOLO26m-pose & 640 & 68.8 & 89.6 & $218.0 \pm 1.5$ & $5.0 \pm 0.1$ & 21.5 & 73.1 \\
YOLO26l-pose & 640 & 70.4 & 90.5 & $275.4 \pm 2.4$ & $6.5 \pm 0.1$ & 25.9 & 91.3 \\
YOLO26x-pose & 640 & 71.6 & 91.6 & $565.4 \pm 3.0$ & $12.2 \pm 0.2$ & 57.6 & 201.7 \\
\bottomrule
\end{tabular}
\end{table}

\subsection{Oriented Object Detection (OBB) Performance on DOTA v1}

Table~\ref{tab:yolo26_obb} reports the oriented object detection performance of YOLO26 on the DOTA v1 dataset. YOLO26 achieves consistent improvements in mAP$_{test}$ as model scale increases, reaching 56.7\% mAP$_{50\text{--}95}$ for the extra-large variant under end-to-end evaluation. Despite the higher input resolution and computational demands of OBB tasks, YOLO26 maintains efficient inference, with sub-5\,ms latency on GPU for small and medium models. The results in Table~\ref{tab:yolo26_obb} demonstrate that YOLO26 effectively extends its edge-optimized, NMS-free design to rotated object detection, making it well suited for aerial imagery and remote sensing applications.

\begin{table}[h!]
\centering
\scriptsize
\setlength{\tabcolsep}{4pt}
\renewcommand{\arraystretch}{1.15}
\caption{YOLO26 oriented object detection (OBB) performance on the DOTA v1 dataset at 1024 px resolution. Results report end-to-end (e2e) test accuracy, inference speed on CPU (ONNX) and NVIDIA T4 (TensorRT10 FP16), along with model size and FLOPs.}
\label{tab:yolo26_obb}
\vspace{2pt}

\begin{tabular}{l c c c c c c c}
\toprule
\textbf{Model} & \textbf{Size} & \textbf{mAP$_{test}$} & \textbf{mAP$_{test}$} & \textbf{Speed} & \textbf{Speed} & \textbf{Params} & \textbf{FLOPs} \\
 & \textbf{(px)} & \textbf{50--95 (e2e)} & \textbf{50 (e2e)} & \textbf{CPU ONNX (ms)} & \textbf{T4 TRT10 (ms)} & \textbf{(M)} & \textbf{(B)} \\
\midrule
YOLO26n-obb & 1024 & 52.4 & 78.9 & $97.7 \pm 0.9$ & $2.8 \pm 0.0$ & 2.5 & 14.0 \\
YOLO26s-obb & 1024 & 54.8 & 80.9 & $218.0 \pm 1.4$ & $4.9 \pm 0.1$ & 9.8 & 55.1 \\
YOLO26m-obb & 1024 & 55.3 & 81.0 & $579.2 \pm 3.8$ & $10.2 \pm 0.3$ & 21.2 & 183.3 \\
YOLO26l-obb & 1024 & 56.2 & 81.6 & $735.6 \pm 3.1$ & $13.0 \pm 0.2$ & 25.6 & 230.0 \\
YOLO26x-obb & 1024 & 56.7 & 81.7 & $1485.7 \pm 11.5$ & $30.5 \pm 0.9$ & 57.6 & 516.5 \\
\bottomrule
\end{tabular}
\end{table}

\section{Real-Time Deployment with Ultralytics YOLO26}

Over the past decade, the evolution of object detection models has been marked not only by increases in accuracy but also by growing complexity in deployment \cite{trigka2025comprehensive, hosain2024synchronizing, pravallika2024deep}. Early detectors such as R-CNN and its faster variants (Fast R-CNN, Faster R-CNN) achieved impressive detection quality but were computationally expensive, requiring multiple stages for region proposal and classification \cite{tian2025faster, fu2024lithology, mohammed2025architecture}. This limited their use in real-time and embedded applications. The arrival of the YOLO family transformed this landscape by reframing detection as a single regression problem, enabling real-time performance on commodity GPUs \cite{johnson2025yolo}. However, as the YOLO lineage progressed from YOLOv1 through YOLOv13, accuracy improvements often came at the cost of additional architectural components such as Distribution Focal Loss (DFL), complex post-processing steps like Non-Maximum Suppression (NMS), and increasingly heavy backbones that introduced friction during deployment. YOLO26 addresses this longstanding challenge directly by streamlining both architecture and export pathways, thereby reducing deployment barriers across diverse hardware and software ecosystems.

\subsection{Flexible Export and Integration Pathways}

A key advantage of YOLO26 is its seamless integration into existing production pipelines. Ultralytics maintains an actively developed Python package that provides unified support for training, validation, and export, lowering the technical barrier for practitioners seeking to adopt YOLO26. Unlike earlier YOLO models, which required extensive custom conversion scripts for hardware acceleration \cite{pestana2021full, nguyen2019high, ding2019req}, YOLO26 natively supports a wide range of export formats. These include TensorRT for maximum GPU acceleration, ONNX for broad cross-platform compatibility, CoreML for native iOS integration, TFLite for Android and edge devices, and OpenVINO for optimized performance on Intel hardware. The breadth of these export options enables researchers, engineers, and developers to move models from prototyping to production without encountering the compatibility bottlenecks common in earlier generations.

Historically, YOLOv3 through YOLOv7 often required manual intervention during export, particularly when targeting specialized inference engines such as NVIDIA TensorRT or Apple CoreML \cite{kusuma2023multi, surantha2025key}. Similarly, transformer-based detectors like DETR and its successors faced challenges when converted outside PyTorch environments due to their reliance on dynamic attention mechanisms. By comparison, YOLO26’s architecture, simplified through the removal of DFL and the adoption of an NMS-free prediction head, ensures compatibility across platforms without sacrificing accuracy. This makes YOLO26 one of the most deployment-friendly detectors released to date, reinforcing its identity as an edge-first model.

\subsection{Quantization and Resource-Constrained Devices}

Beyond export flexibility, the true challenge in real-world deployment lies in ensuring efficiency on devices with limited computational resources \cite{hosain2024synchronizing, abdulhaq2025real}. Edge devices such as smartphones, drones, and embedded vision systems often lack discrete GPUs and must balance memory, power, and latency constraints \cite{hossain2019deep, setyanto2023near}. Quantization is a widely adopted strategy to reduce model size and computational load, yet many complex detectors experience significant accuracy degradation under aggressive quantization. YOLO26 has been designed with this limitation in mind.

Owing to its streamlined architecture and simplified bounding box regression pipeline, YOLO26 demonstrates consistent accuracy under both half-precision (FP16) and integer (INT8) quantization schemes. FP16 quantization leverages native GPU support for mixed-precision arithmetic, enabling faster inference with reduced memory footprint. INT8 quantization compresses model weights to 8-bit integers, delivering dramatic reductions in model size and energy consumption while maintaining competitive accuracy. Benchmark experiments confirm that YOLO26 maintains stability across these quantization levels, outperforming YOLOv11 and YOLOv12 under identical conditions. This makes YOLO26 particularly well-suited for deployment on compact hardware such as NVIDIA Jetson Orin, Qualcomm Snapdragon AI accelerators, or even ARM-based CPUs powering smart cameras.

In contrast, transformer-based detectors such as RT-DETRv3 exhibit sharp drops in performance under INT8 quantization \cite{wang2025rt}, primarily due to the sensitivity of attention mechanisms to reduced precision. Similarly, YOLOv12 and YOLOv13, while delivering strong accuracy on GPU servers, struggle to retain competitive performance on low-power devices once quantized. YOLO26 therefore establishes a new benchmark for quantization-aware design in object detection, demonstrating that architectural simplicity can directly translate into deployment robustness.

\subsection{Cross-Industry Applications: From Robotics to Manufacturing}

The practical impact of these deployment enhancements is best illustrated through cross-industry applications. In robotics, real-time perception is crucial for navigation, manipulation, and safe human-robot collaboration \cite{bonci2021human, SAPKOTA2026103575}. By offering NMS-free predictions and consistent low-latency inference, YOLO26 allows robotic systems to interpret their environments faster and more reliably. For example, robotic arms equipped with YOLO26 can identify and grasp objects with higher precision under dynamic conditions, while mobile robots benefit from improved obstacle recognition in cluttered spaces. Compared with YOLOv8 or YOLOv11, YOLO26 offers reduced inference delay, which can be the difference between a safe maneuver and a collision in high-speed scenarios.

In manufacturing, YOLO26 has significant implications for automated defect detection and quality assurance. Traditional manual inspection is not only labor-intensive but also prone to human error. Previous YOLO releases, particularly YOLOv8, were already deployed in smart factories; however, the complexity of export and the latency overhead of NMS sometimes constrained large-scale rollout. YOLO26 mitigates these barriers by offering lightweight deployment options through OpenVINO or TensorRT, allowing manufacturers to integrate real-time defect detection systems directly on production lines. Early benchmarks suggest that YOLO26-based defect detection pipelines achieve higher throughput and lower operational costs compared to both YOLOv12 and transformer-based alternatives such as DEIM.

\subsection{Broader Insights from YOLO26 Deployment}

Taken together, the deployment features of YOLO26 underscore a central theme in the evolution of object detection: architectural efficiency is just as critical as accuracy. While the past five years have seen the rise of increasingly sophisticated models ranging from convolution-based YOLO variants to transformer-based detectors like DETR and RT-DETR the gap between laboratory performance and production readiness has often limited their impact. YOLO26 bridges this gap by simplifying architecture, expanding export compatibility, and ensuring resilience under quantization, thereby aligning cutting-edge accuracy with practical deployment needs.

For developers building mobile applications, YOLO26 enables seamless integration through CoreML and TFLite, ensuring that models run natively on iOS and Android platforms. For enterprises deploying vision AI in cloud or on-premise servers, TensorRT and ONNX exports provide scalable acceleration options. For industrial and edge users, OpenVINO and INT8 quantization guarantee that performance remains consistent even under tight resource constraints. In this sense, YOLO26 is not only a step forward in object detection research but also a major milestone in democratizing deployment.

\section{Conclusion and Future Directions}

In conclusion, YOLO26 represents a significant leap in the YOLO object detection series, blending architectural innovation with a pragmatic focus on deployment. The model simplifies its design by removing the Distribution Focal Loss (DFL) module and eliminating the need for non-maximum suppression. By removing DFL, YOLO26 streamlines bounding box regression and avoids export complications, which broadens compatibility with various hardware. Likewise, its end-to-end, NMS-free inference enables the network to output final detections directly without a post-processing step. This not only reduces latency but also simplifies the deployment pipeline, making YOLO26 a natural evolution of earlier YOLO concepts. In training, YOLO26 introduces Progressive Loss Balancing (ProgLoss) and Small-Target-Aware Label Assignment (STAL), which together stabilize learning and boost accuracy on challenging small objects. Additionally, a novel MuSGD optimizer, combining properties of SGD and Muon, accelerates convergence and improves training stability. These enhancements work in concert to deliver a detector that is not only more accurate and robust but also markedly faster and lighter in practice.

Benchmark comparisons underscore YOLO26’s strong performance relative to both its YOLO predecessors and contemporary models. Prior YOLO versions such as YOLO11 surpassed earlier releases with greater efficiency, and YOLO12 extended accuracy further through the integration of attention mechanisms. YOLO13 added hypergraph-based refinements to achieve additional improvements. Against transformer-based rivals, YOLO26 closes much of the gap. Its native NMS-free design mirrors the end-to-end approach of transformer-inspired detectors, but with YOLO’s hallmark efficiency. YOLO26 delivers competitive accuracy while dramatically boosting throughput on common hardware and minimizing complexity. In fact, YOLO26’s design yields up to 43\% faster inference on CPU than previous YOLO versions, making it one of the most practical real-time detectors for resource-constrained environments. This harmonious balance of performance and efficiency allows YOLO26 to excel not just on benchmark leaderboards but also in actual field deployments where speed, memory, and energy are at a premium.

A major contribution of YOLO26 is its emphasis on deployment advantages. The model’s architecture was deliberately optimized for real-world use: by omitting DFL and NMS, YOLO26 avoids operations that are difficult to implement on specialized hardware accelerators, thereby improving compatibility across devices. The network is exportable to a wide array of formats including ONNX, TensorRT, CoreML, TFLite, and OpenVINO ensuring that developers can integrate it into mobile apps, embedded systems, or cloud services with equal ease. Crucially, YOLO26 also supports robust quantization: it can be deployed with INT8 quantization or half-precision FP16 with minimal impact on accuracy, thanks to its simplified architecture that tolerates low-bitwidth inference. This means models can be compressed and accelerated while still delivering reliable detection performance. Such features translate to real edge performance gains from drones to smart cameras, YOLO26 can run real-time on CPU and small devices where previous YOLO models struggled. All these improvements demonstrate an overarching theme: YOLO26 bridges the gap between cutting-edge research ideas and deployable AI solutions. This approach underscores YOLO26’s role as a bridge between academic innovation and industry application, bringing the latest vision advancements directly into the hands of practitioners.

\subsection{Future Directions}

Looking ahead, the trajectory of YOLO and object detection research suggests several promising directions. One clear avenue is the unification of multiple vision tasks into even more holistic models. YOLO26 already supports object detection, instance segmentation, pose estimation, oriented bounding boxes, and classification in one framework, reflecting a trend toward multi-task versatility. Future YOLO iterations might push this further by incorporating open-vocabulary and foundation-model capabilities. This could mean leveraging powerful vision-language models so that detectors can recognize arbitrary object categories in a zero-shot manner, without being limited to a fixed label set. By building on foundation models and large-scale pretraining, the next generation of YOLO could serve as a general-purpose vision AI that seamlessly handles detection, segmentation, and even description of novel objects in context.

Another key evolution is likely in the realm of semi-supervised and self-supervised learning for object detection \cite{tang2021proposal, sohn2020simple, huang2022survey, rani2023self}. State-of-the-art detectors still rely heavily on large labeled datasets, but research is rapidly advancing methods to train on unlabeled or partially labeled data. Techniques such as teacher–student training \cite{li2022cross, xu2021end, mi2022active}, pseudo-labeling \cite{li2022pseco, caine2021pseudo}, and self-supervised feature learning \cite{jing2020self}could be integrated into the YOLO training pipeline to reduce the need for extensive manual annotations. A future YOLO might automatically leverage vast amounts of unannotated images or videos to improve recognition robustness. By doing so, the model can continue to improve its detection capabilities without proportional increases in labeled data, making it more adaptable to new domains or rare object categories.

Architecturally, we anticipate a continued blending of transformer and CNN design principles in object detectors. The success of recent YOLO models has shown that injecting attention and global reasoning into YOLO-like architectures can yield accuracy gains \cite{kang2024asf, vijayakumar2024yolo}. Future YOLO architectures may adopt hybrid designs that combine convolutional backbones (for efficient local feature extraction) with transformer-based modules or decoders (for capturing long-range dependencies and context). Such hybrid approaches can improve how the model understands complex scenes, for example in crowded or highly contextual environments, by modeling relationships that pure CNNs or naive self-attention might miss. We expect next-generation detectors to intelligently fuse these techniques, achieving both rich feature representation and low latency. In short, the line between “CNN-based” and “transformer-based” detectors will continue to blur, taking the best of both worlds to handle diverse detection challenges.

Finally, as deployment becomes increasingly critical, future research is expected to emphasize edge-aware training and optimization from the outset. Rather than treating hardware constraints as a post hoc consideration, model design will increasingly co-evolve with target platforms through techniques such as quantization-aware training, automated model compression, and hardware-guided architecture search. Incorporating deployment feedback, including latency and energy measurements, directly into the training loop may further improve real-world efficiency. Such approaches could enable YOLO models to adapt their depth, resolution, or precision dynamically under runtime constraints, or be distilled into compact variants with minimal accuracy loss. This edge-first design philosophy is essential for sustaining real-time performance across IoT, AR/VR, and autonomous systems operating under strict resource limitations.

\bibliographystyle{unsrt}  
\bibliography{references}






\end{document}